\documentclass[sigconf,authorversion,nonacm]{acmart}

\usepackage{algorithm}
\usepackage{algorithmic}
\usepackage{multirow}
\usepackage{listings}
\usepackage{enumitem}

\def\etal{\textit{et al}.}

\AtBeginDocument{%
  }

\acmConference[ ]{Make sure to enter the correct conference title from your rights confirmation email}{Oct 27, 2025}{Dublin, Ireland}
\acmYear{} 
\acmBooktitle{}
\acmISBN{}
\acmSubmissionID{}

\begin{document}

\title{Robustness in AI-Generated Detection: Enhancing Resistance to Adversarial Attacks}

\author{Haoxuan Sun} 
\author{Jiahui Zhan}
\email{guwangtu@sjtu.edu.cn} 
\affiliation{%
  \institution{Shanghai Jiao Tong University}
  \city{Shanghai}
  \country{China}
} 

\author{Yan Hong}
\author{Haoxing Chen}
\author{Jun Lan}
\author{Huijia Zhu}
\author{Weiqiang Wang}
\affiliation{%
  \institution{Alibaba Group}
  \city{Shanghai}
  \country{China}} 

\author{Liqing Zhang}
\author{Jianfu Zhang}
\affiliation{%
  \institution{Shanghai Jiao Tong University}
  \city{Shanghai}
  \country{China}}
\email{c.sis@sjtu.edu.cn}
\renewcommand\footnotetextcopyrightpermission[1]{}

\settopmatter{printacmref=false} 

\begin{abstract}
The rapid advancement of generative image technology has introduced significant security concerns, particularly in the domain of face generation detection. This paper investigates the vulnerabilities of current AI-generated face detection systems. Our study reveals that while existing detection methods often achieve high accuracy under standard conditions, they exhibit limited robustness against adversarial attacks. To address these challenges, we propose an approach that integrates adversarial training to mitigate the impact of adversarial examples. Furthermore, we utilize diffusion inversion and reconstruction to further enhance detection robustness. Experimental results demonstrate that minor adversarial perturbations can easily bypass existing detection systems, but our method significantly improves the robustness of these systems. Additionally, we provide an in-depth analysis of adversarial and benign examples, offering insights into the intrinsic characteristics of AI-generated content. All associated code will be made publicly available in a dedicated repository to facilitate further research and verification.
\end{abstract}

 




\maketitle

\section{Introduction}
The rapid progress of generative models ~\cite{gan,cyclegan,stylegan}, particularly diffusion models ~\cite{ddpm,dmbgi,htigcl,ldm,dit,var}, has significantly improved the authenticity of face generation technology.  
These methods have enabled the creation of highly realistic images that can often bypass detection, even by human observers. 
While this progress demonstrates the remarkable potential of artificial intelligence in image generation, it also introduces a major concern: the misuse of such high-fidelity synthetic images could contribute to the spread of misinformation and fake content. 
The ability to generate such high-quality images raises ethical issues and underscores the urgent need for robust detection systems to safeguard the integrity of digital media.

\begin{figure}[t]
    \centering
    \includegraphics[width=1\linewidth]{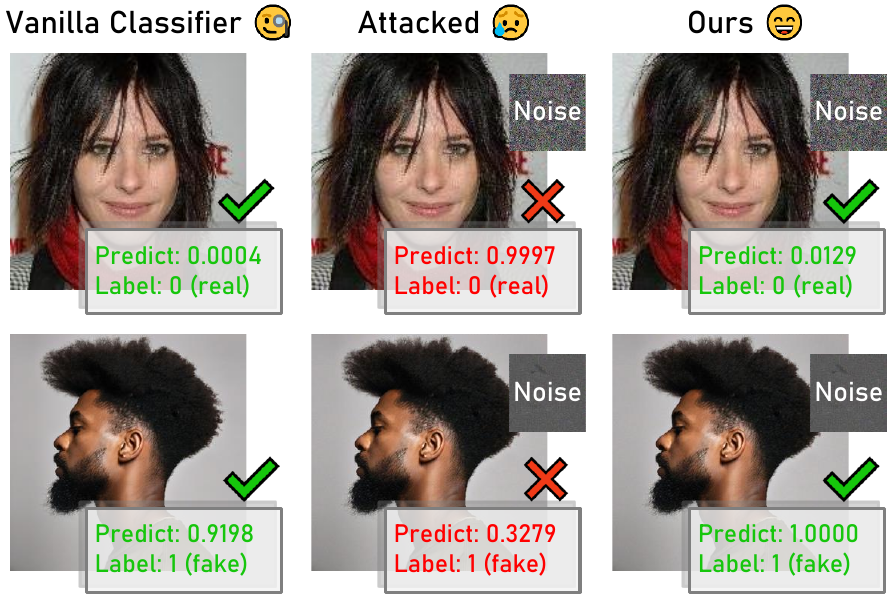}
    \caption{Although SOTA detectors achieve near-perfect accuracy in discriminating fake images, even minimal and imperceptible perturbations can cause misclassifications, leading to synthetic images being identified as real, and vice versa. In contrast, our proposed method maintains robust classification accuracy even under adversarial attacks.}
    \label{fig:motivation}
\end{figure}

Recent strides in AIGC (Artificial Intelligence Generated Content) detection~\cite{dire,Lasted,drct,code} have achieved remarkable accuracy under standard conditions, with most state-of-the-art generated images classified from real images with near-perfect accuracy.
However, a critical vulnerability persists: these models often fail when subjected to adversarial attacks ~\cite{fgm,fgsm,pgd,autoattack}. 
Our research reveals that similar to classification tasks, even minimal and imperceptible perturbations can lead to misclassifications, causing detection systems to incorrectly identify synthetic images as real and vice versa as shown in Fig.~\ref{fig:motivation}. 
Although there have been many studies on adversarial robustness in current research, there is a lack of research on the robustness of AIGC detection systems, underscoring a critical area for improvement in detection system defenses. Current research lacks advanced techniques capable of adapting to and mitigating the detrimental effects of adversarial attacks. This remains a pressing issue, as adversarial perturbations are imperceptible to humans but can almost completely disrupt detection models.

To support the practical deployment of more stable and secure detection systems in real-world scenarios, we analyze two settings of AI-generated image detection: in-domain settings and cross-domain settings. We evaluate both detection performance and robustness under adversarial attacks. Our findings indicate that adversarial training effectively mitigates the impact of adversarial perturbations. Furthermore, DIffusion Reconstruction Error~(DIRE)~\cite{dire}, the inverse reconstruction residue of the diffusion models, further enhances robustness. As shown in Fig.~\ref{fig:motivation}, our new strategy significantly improves the detector's resistance.
In summary, this paper makes the following contributions:
\begin{itemize}[leftmargin=*]
    \item Experiments reveal that even the most advanced AI-generated face detection models are easily deceived by minor adversarial perturbations. We are the first to focus on this critical issue in AIGC detection. 
    \item We demonstrate that adversarial training and diffusion reconstruction error enhance the robustness of detection models when the test samples and training samples follow the same distribution. 
    \item Through a detailed performance analysis across different datasets, we show that adversarial training alone struggles to generalize to new, unseen data distributions. Incorporating the diffusion reconstruction error approach significantly improves adversarial generalizability. 
\end{itemize}

\section {Related Works}

AIGC (\textit{i.e.}, Deepfakes) refers to images, videos, or audio that are edited or generated using artificial intelligence. AI-generated content detection aims to differentiate deepfakes from natural images, videos, or audio. 
Early detection methods focused on identifying artifacts and inconsistencies caused by the limitations of GANs~\cite{gan,cyclegan,karras2019style}, such as anomalies in lighting~\cite{186,187}, shadows~\cite{185}, and reflections~\cite{185}. However, as models such as diffusion models~\cite{ddpm,iddpm,dmbgi,htigcl,ldm}, DiT~\cite{dit} and VAR~\cite{var} improved realism and reduced artifacts, these techniques became less effective. 

Recent research emphasizes unique traces left by image generation processes. Deep Image Fingerprint (DIF)~\cite{102} uses convolutional neural networks to extract unique fingerprints for identifying images from specific models. 
DIRE~\cite{dire} and SeDID~\cite{184} utilize reverse diffusion to detect subtle differences between real and synthetic images. DIRE focuses on reconstruction accuracy at the initial timestep, while SeDID leverages errors from intermediate diffusion steps for richer analysis. 
Diffusion Reconstruction Contrastive Learning (DRCT)~\cite{drct} generates challenging samples through high-quality diffusion reconstruction and employs contrastive training to improve the detector's generalizability. 
Addressing dataset imbalances, Xu \etal ~\cite{xu2024analyzing} highlighted how attribute inconsistencies negatively affect detection results and proposed the creation of annotated datasets to remedy this issue. 
Furthermore, the Contrastive Deepfake Embeddings (CoDE) ~\cite{code} framework introduces a new embedding space trained through contrastive learning that emphasizes global-local similarities to enhance detection performance. To combat generalization issues stemming from overfitting to specific artifacts, 
Latent Space Data Augmentation (LSDA)~\cite{lsda} expands forgery representation diversity, thereby enabling models to adopt more flexible decision boundaries. 
Moreover, Tan \etal ~\cite{rethinking} focus on generator architectures, specifically rethinking CNN-based structures to show how upsampling operators can produce generalized forgery artifacts and introducing Neighboring Pixel Relationships (NPR) for effective characterization of these artifacts, leading to notable performance gains across various datasets. 
Additionally, while some studies have concentrated on video deepfake detection~\cite{v1,v2,v3}, this paper focuses on still image detection and will not elaborate on those contributions.
Incorporating multimodal information has also shown promise. Lasted~\cite{Lasted} uses language-guided contrastive learning with text labels like  ``real/synthetic photo'' and ``real/synthetic painting''  to uncover forensic features and improve detection across diverse generative methods. 

Despite these advancements, existing detection methods remain vulnerable to adversarial attacks. Although adversarial perturbations are often imperceptible to humans, they can significantly degrade the performance of detection models.
Techniques like Fast Gradient Sign Method (FGSM)~\cite{fgsm}, Fast Gradient Method (FGM)~\cite{fgm}, and Projected Gradient Descent (PGD)~\cite{pgd} generate adversarial examples by altering inputs based on loss gradients. 
Among these, PGD, with its iterative updates and constraint projections, is particularly effective at producing adversarial samples.
In the field of facial recognition, there have been many studies on improving model robustness~\cite{advdefence1,advdefence2,advdefence3}, but there is still a gap in research on AIGC detection. Furthermore, some works utilize generative models to defend against attacks~\cite{review1_4,review1_5}. However, for the task of distinguishing whether an image is generated by AI, we should avoid using generative models in the judgment process to prevent artifacts from the generative models from interfering with the assessment.

Our study shows that even subtle perturbations can severely impact the performance of current frameworks, emphasizing the need for robust detection systems that can withstand adversarial conditions and reliably identify AI-generated images.

\begin{figure*}[t]
    \centering
    \includegraphics[width=0.95\linewidth]{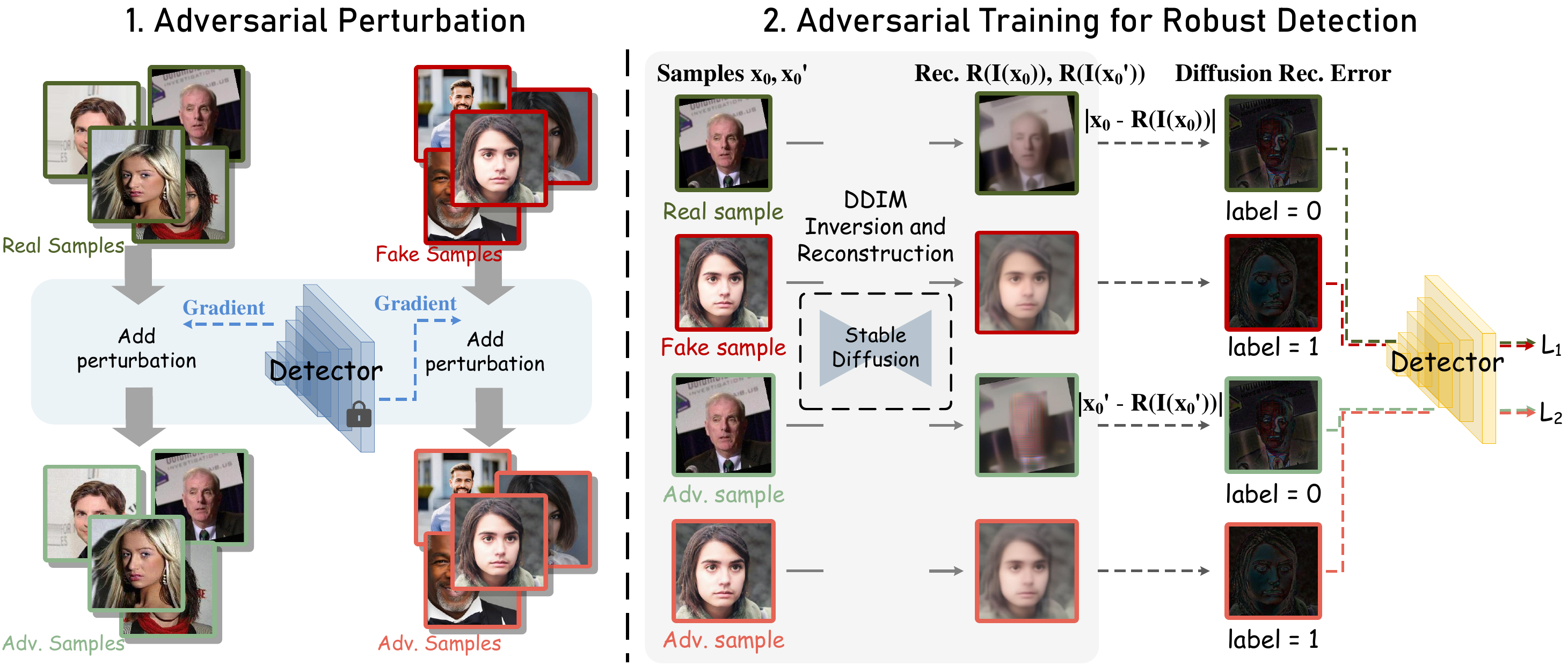}
    \caption{Pipeline of our robustness-centric method, which integrates adversarial training with diffusion inversion reconstruction to improve detection robustness. Given a set of real and fake images, we first apply adversarial perturbations and then use the resulting adversarial samples along with the original samples for residual reconstruction. Finally, we train the detector using the reconstruction residual maps for more robust performance.}
    \label{fig:pipeline}
\end{figure*}

\begin{algorithm}[t]
\caption{Adversarial Training with DIRE}
\begin{algorithmic}[1]
\label{alg:adv_dire}
\STATE \textbf{Input:} Detect Model $f$, dataloader $D$, number of epochs $E$, step size $\alpha$, perturbation bound $\varepsilon$, number of PGD iterations $T$.
  
\FOR{epoch = 1 to $E$}
    \FOR{(x, y) in $D$}
        \STATE $x' \leftarrow x +  \delta_0$   
        \FOR{t = 1 to $T$}
            \STATE $x' \leftarrow x' + \alpha \cdot \text{sign}(\nabla_{x'} \mathcal{L}(f(x'), y))$   
            \STATE $x' \leftarrow \text{B}(x', x - \varepsilon, x + \varepsilon)$   
        \ENDFOR
        \STATE $x_0 \leftarrow \textrm{DDIM Reconstruction}(\textrm{DDIM Inversion}(x))$
        \STATE $x_0' \leftarrow \textrm{DDIM Reconstuction}(\textrm{DDIM Inversion}(x'))$
        \STATE $ \textit{DIRE}(x_0) \leftarrow |x_0-x|$
        \STATE $ \textit{DIRE}(x_0') \leftarrow |x_0'-x'|$
        \STATE $x_{\text{combined}} \leftarrow \text{Concat}(\textit{DIRE}(x_0), \textit{DIRE}(x_0'))$
        \STATE $y_{\text{combined}} \leftarrow \text{Concat}(y, y)$  
        \STATE $y_{\text{predicted}} \leftarrow f(x_{\text{combined}})$  
        \STATE $\mathcal{L}_{total} \leftarrow \mathcal{L}(y_{\text{predicted}}, y_{\text{combined}})$  
        \STATE $f \leftarrow \text{Update}(\mathcal{L}_{total},f)$
    \ENDFOR
\ENDFOR
\STATE \textbf{Output:} Trained model $f$
\end{algorithmic}
\end{algorithm}

\section{Methodologies}


Given an input image $x$, we define $f(\cdot)$ as the detection model. 
The function $f(x)$ outputs a binary classification label, determining whether the input image is real or synthetic. 
Each image is assigned a label $y\in \{0, 1\}$ where 1 indicates a synthetic image and 0 indicates a real image. The goal of $f(\cdot)$ is to map $f(x)$ to the corresponding $y$.
Recent studies~\cite{drct, code} have demonstrated that simple classification models achieve near-perfect accuracy under standard conditions, successfully distinguishing most of the state-of-the-art generated images from real ones. However, we will show that these models remain highly susceptible to adversarial perturbations.
Our complete pipeline for enhancing the robustness of the detection model is illustrated in Fig.\ref{fig:pipeline} and detailed in Algorithm\ref{alg:adv_dire}.

\subsection{Adversarial Perturbation}
This section outlines the process of generating adversarial examples against AI-generated image detection models using the Projected Gradient Descent (PGD)~\cite{pgd} method, a widely adopted approach to craft adversarial perturbations capable of deceiving machine learning models.
The PGD method begins with a clean input image, denoted as $x$. The objective is to generate an adversarial example $x'$ by adding a small perturbation $\delta $ to $x$. 
This perturbation is designed to be imperceptible to the human eye while effectively misleading the model into making incorrect predictions.

\paragraph{Initialization} 

The process starts by initializing the perturbation $\delta $, where $\delta $ is uniformly distributed within the $\varepsilon$ neighborhood, where $\varepsilon$ defines the maximum allowable perturbation, resulting in the first adversarial example as $x'=x+\delta_0$.

\paragraph{Gradient Calculation}

The next step involves calculating the gradient of the loss function with respect to the input image. This is performed by conducting a forward pass through the model to obtain the predicted output $f(x)$ and then computing the gradient:
\begin{equation}
    g=\Delta_x  \mathcal{L}(f(x),y),
\end{equation}
where $\mathcal{L}$ is the loss function, $f(x)$ is the output of model given $x$ as input, and $y$ is the true label.

\paragraph{Update the Perturbation}

The perturbation is then updated in the direction of the gradient to maximize the loss:
\begin{equation}
    \delta_{i+1}=\delta_i+\alpha \cdot \textrm{sign}(g),
\end{equation}
where $\alpha$ is a small step size that controls the magnitude of the perturbation.

\paragraph{Bound}
To ensure that the adversarial perturbation remains imperceptible, a bound is applied to constrain its magnitude. Specifically, the perturbation is restricted within a predefined norm $L_p$ and is projected as:
\begin{equation}
    \delta=B(\delta,-\varepsilon,\varepsilon),
\end{equation}

\paragraph{Iteration}

The steps \textit{gradient calculation} through the \textit{bound} are repeated for a predefined number of iterations, refining the perturbation each time to produce the final adversarial example $x'=x+\delta$.

Finally, the adversarial example $x'_n$ after $n$ iterations can be expressed as:
\begin{equation}
        x'_0= x + \delta_0,
        x'_n= B_{x',\varepsilon}(x'_{n-1}+\alpha\cdot \mathrm{sign}(\Delta_x \mathcal{L}(x'_{n-1},y))),
\end{equation}




The effectiveness of these adversarial perturbations lies in their ability to exploit the inherent vulnerabilities of machine learning models. Despite being nearly invisible to the human eye, these perturbations can induce significant changes in model predictions. For example, a facial recognition system might incorrectly classify the adversarial example as a different individual or fail to detect a face entirely. 
Similarly, adding such perturbations to real images results in their misclassification as AI-generated images, while perturbing generated images leads to their misclassification as real images. 
\textit{Due to the intrinsic characteristics of the generated images, this type of noise is even more difficult to detect, as it is challenging to discern whether the noise originates from the adversarial attack or from the generative model itself. }
This phenomenon highlights the critical security risks of deploying deep learning models in sensitive applications, where even minor, undetectable modifications can severely compromise their accuracy and reliability.



\begin{table*}[t]
    \centering
    \caption{Performance obtained through training on all five datasets using different methods.}
    \label{tab:single_model}

    \begin{tabular}{@{}ccccccccc@{}}
        \toprule
        \multirow{2}{*}{Method} & \multirow{2}{*}{Dataset} & \multicolumn{2}{c}{Clean Images} & \multicolumn{2}{c}{Adversarial Images} &  \multicolumn{2}{c}{Robustness Score} \\ \cmidrule(lr){3-4} \cmidrule(lr){5-6} \cmidrule(lr){7-8}
                                  &                        &w/o AT &w/ AT   & w/o AT & w/ AT   & w/o AT & w/ AT  \\ \midrule
        \multirow{5}{*}{ResNet}  & CelebA                   & 99.99\%              & 99.99\%             & 0.00\%             & 99.02\%          & 0.00      & 0.99                \\ 
                                  & LFW                      & 99.97\%              & 99.92\%             & 0.00\%             & 89.02\%           & 0.00     & 0.89                \\ 
                                  & Selfie                   & 99.99\%              & 99.72\%             & 0.00\%             & 97.18\%           & 0.00      & 0.97                \\ 
                                  & SFHQ                     & 99.99\%              & 99.99\%             & 0.00\%             & 99.98\%           &  0.00   & 0.99                \\ 
                                  & SDFace                   & 99.98\%              & 82.53\%             & 0.00\%             & 70.77\%           &  0.00   & 0.85                \\ \midrule
        \multirow{5}{*}{ViT}     & CelebA                   & 99.99\%              & 99.96\%             & 0.00\%             & 99.84\%            & 0.00   & 0.99               \\ 
                                  & LFW                      & 99.99\%              & 99.50\%             & 0.00\%             & 99.01\%           &  0.00   & 0.99                \\ 
                                  & Selfie                   & 99.99\%              & 98.89\%             & 0.00\%             & 97.77\%           & 0.00    & 0.99                \\ 
                                  & SFHQ                     & 99.99\%              & 99.98\%             & 0.00\%             & 99.98\%           &  0.00   & \textbf{1.00}                \\ 
                                  & SDFace                   & 99.85\%              & 85.02\%             & 0.00\%             & 75.78\%           &  0.00   & 0.89                \\ \midrule
        \multirow{5}{*}{DIRE}    & CelebA                   & 100.0\%              & 100.0\%             & 81.24\%             & \textbf{99.99\%}            &  0.81  & \textbf{1.00}            \\ 
                                  & LFW                      & 100.0\%              & 100.0\%             & 97.66\%             & \textbf{100.0\%}           & 0.97    & \textbf{1.00}                \\ 
                                  & Selfie                   & 99.89\%              & 99.77\%             & 98.07\%             & \textbf{99.91\%}          & 0.98    & \textbf{1.00}                \\ 
                                  & SFHQ                     & 100.0\%              & 100.0\%             & 99.73\%             & \textbf{99.99\%}           &  0.99   & \textbf{1.00}                \\ 
                                  & SDFace                   & 99.89\%              & 99.89\%             & 38.94\%             & \textbf{99.89\%}            & 0.39   & \textbf{1.00}               \\ 
       
        \bottomrule
    \end{tabular}
\end{table*}

\begin{figure}[t]
    \centering
    \includegraphics[width=1\linewidth]{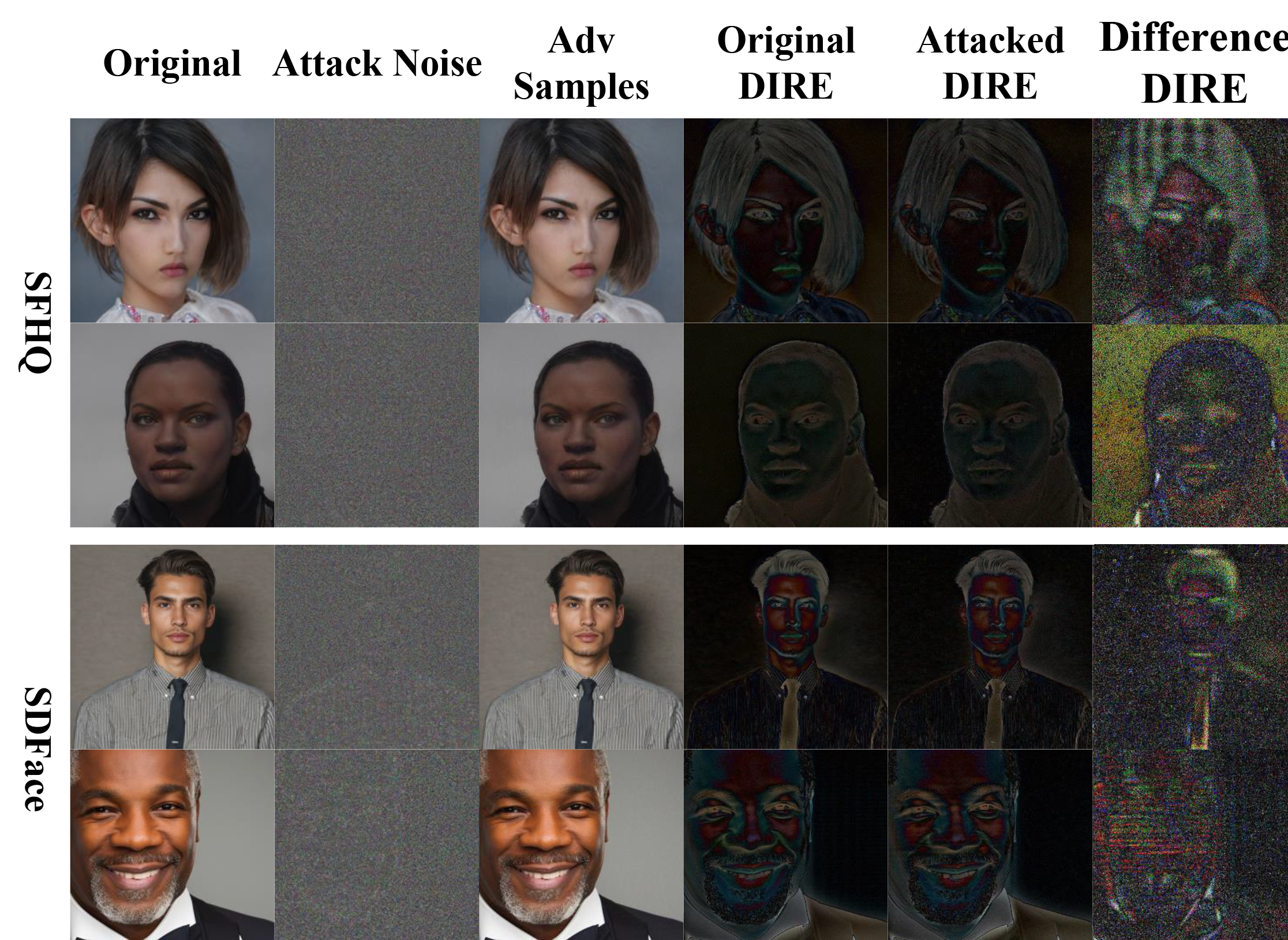} 
    \caption{
    Visualization of attack noise, adversarial samples, and DIRE residual maps of \textbf{fake images}.
``Difference DIRE” represents the disparity between DIRE maps of the original samples and those of the adversarially attacked samples. For enhanced clarity, the attack noise was amplified by a factor of 20, and the Difference DIRE maps by a factor of 10.
    }
    \label{fig:cp}
\end{figure}

\subsection{Adversarial Training}
One effective approach to defending against adversarial perturbations is adversarial training. Adversarial training~\cite{pgd} is a widely used defense mechanism designed to improve the robustness of machine learning models against adversarial attacks. By incorporating adversarial examples into the training process, the model learns to adapt to perturbations that would otherwise lead to misclassifications, thereby enhancing its resistance to adversarial manipulation.

The training process uses a combined loss function that accounts for both the standard loss in clean examples and the loss in adversarial examples. This can be formulated as:
\begin{equation}
    \label{equ:loss}
    \mathcal{L}_{total}=\mathcal{L}_1(f(x),y)+\lambda\cdot\mathcal{L}_2(f(x'),y),
\end{equation}
where, $\mathcal{L}_1$ and $\mathcal{L}_2$ are cross-entropy loss, $f(x)$ represents the model's prediction, $\lambda$ is a hyperparameter balancing the contributions of clean and adversarial losses, $x'$ is the adversarial example generated using the method described earlier.


The training process uses mixed data, each batch containing clean images and their corresponding adversarial examples. This ensures that the model learns not only the standard features of the data but also the features susceptible to adversarial exploitation. By exposing the model to these challenging examples, its decision boundaries become more robust against perturbations.
Incorporating adversarial examples into the training pipeline significantly enhances the robustness of the model. Our subsequent experiments also confirm that adversarial training effectively improves robustness in AI-generated content detection. By learning from adversarial perturbations, the model develops a deeper understanding of the input space, enabling it to maintain accurate predictions even under adversarial manipulations.

\begin{figure}[t]
    \centering
    \includegraphics[width=1\linewidth]{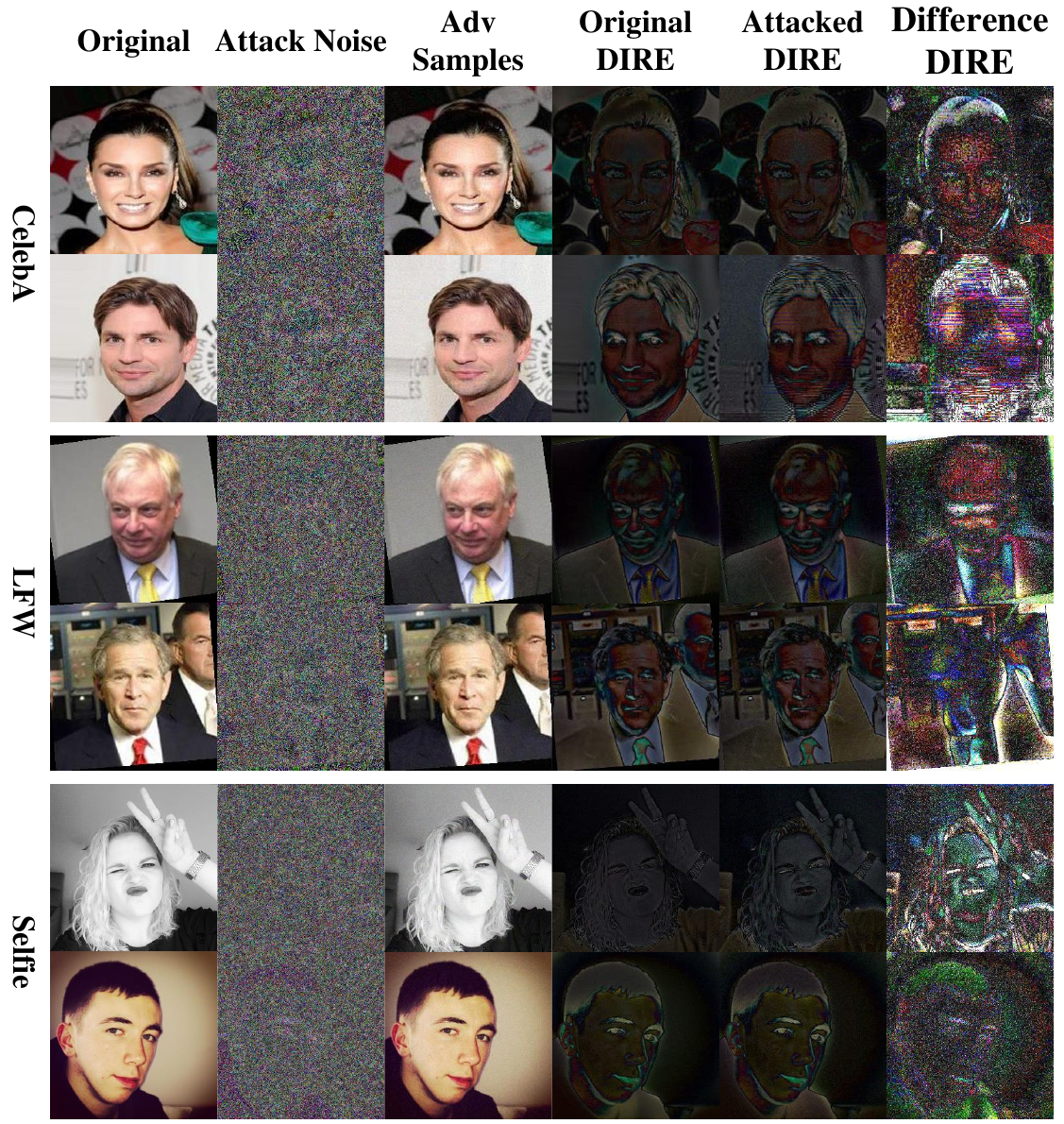} 
    \caption{Visualization of attack noise, adversarial samples, and DIRE maps of \textbf{real images}.}
    \label{fig:cp1}
\end{figure}


\subsection{Diffusion Reconstruction for Robustness Enhancement}
Although adversarial training significantly mitigates the impact of adversarial attacks, we found that when training data is insufficient, detection models remain vulnerable to attacks on data distributions different from the training set, even when adversarial examples are incorporated. 
To address this limitation, we leverage the robustness of the diffusion reconstruction against white-box attacks and propose enhancements to further improve its resilience.

In diffusion models~\cite{ddpm}, the forward process transforms sample $x_0$ to noise latent $x_T$ by progressively adding Gaussian noise, while the reverse process denoises $x_T$ back to $x_0$. $T$ is the number of steps.  
DIffusion Reconstruction Error~(DIRE)~\cite{dire} employs the DDIM~\cite{ddim} inversion process to gradually add noise to $x_0$, mapping it into a noise latent space. 
The reverse process of diffusion models in DDIM is defined as:
\begin{equation}
    \begin{aligned}
        x_{t-1} &= \sqrt{\alpha_{t-1}}\left(\frac{x_t-\sqrt{1-\alpha_t}\epsilon_\theta(x_t,t)}{\sqrt{\alpha_t}}\right) \\
        &\quad +\sqrt{1-\alpha_{t-1}-\sigma_t^2}\cdot\epsilon_\theta(x_t,t)+\sigma_t\epsilon_t,
    \end{aligned} \label{eq:reconstruct}
\end{equation}
where $\epsilon_\theta(x_t,t)$ represents the noise calculated by the noise prediction model given $x_t$ and $t$, and its model parameter is $\theta$. 
$\sigma_t$ represents the standard deviation parameter in the time step $t$, which decreases as $T$ increases.
If $\sigma_t =0$ ($T$ is large enough) the reverse process becomes deterministic (\textit{reconstruction process}), in which one noise latent $x_T$ determines one generated sample $x_0$.

The DDIM inversion process deterministically maps $x_0$ to $x_T$, which can be treated as the reversion of the reconstruction process in Eq.~\ref{eq:reconstruct}:
\begin{equation}
    \frac{x_{t+1}}{\sqrt{\alpha_{t+1}}}=\frac{x_{t}}{\sqrt{\alpha_{t}}}+\left(\sqrt{\frac{1-\alpha_{t+1}}{\alpha_{t+1}}}-\sqrt{\frac{1-\alpha_{t}}{\alpha_{t}}}\right)\epsilon_\theta(x_t,t),
\end{equation}
After $T$ steps, $x_0$ becomes a point $x_T$ in the isotropic Gaussian noise distribution. The inversion process identifies the corresponding point in the noisy space, and the reconstruction process is then used to reconstruct the input image, producing a reconstructed version $x_0'$. 



The differences between $x_0$ and $x_0'$ help to distinguish real or generated. The DIRE map is then defined as:
\begin{equation}
    \textit{DIRE}(x_0)=|x_0-R(I(x_0))|,
\end{equation}
where $|\cdot|$ denotes the computation of the absolute value, and $I(\cdot)$ is a series of the inversion process. $R(\cdot)$ is a series of the reconstruction process.  
Images generated by diffusion models are sampled from the distribution of the diffusion generation space, whereas real images originate from a distinct distribution. As a result, samples from the diffusion generation space are more likely to be faithfully reconstructed by a pre-trained diffusion model, while real images are not.
Therefore, DIRE naturally benefits AI-generated content detection. Additionally, the multi-step inversion and reconstruction process can mitigate the impact of adversarial perturbations.
Experimental results demonstrate that DIRE exhibits a certain level of resistance to adversarial attacks, even when directly confronted with them. More importantly, while adversarial training struggles to handle adversarial examples from different distributions (cross-dataset scenarios) when training data is limited, incorporating DIRE significantly enhances adversarial robustness in such cases. We will further validate and analyze this improvement through subsequent experiments.

\begin{table*}[ht]
    \centering
    \caption{Performance obtained through training on LFW and SFHQ using different methods.}
    \label{tab:cross-model_lfw_sfhq}

    \begin{tabular}{@{}ccccccccc@{}}
        \toprule
        \multirow{2}{*}{Method} & \multirow{2}{*}{Dataset} & \multicolumn{2}{c}{Clean Images} & \multicolumn{2}{c}{Adversarial Images} &  \multicolumn{2}{c}{Robustness Score} \\ \cmidrule(lr){3-4} \cmidrule(lr){5-6} \cmidrule(lr){7-8}
                                  &                           & w/o AT     & w/ AT             & w/o AT        & w/ AT   & w/o AT & w/ AT  \\ \midrule
        \multirow{5}{*}{ResNet}  & CelebA                   & 94.35\%              & 1.58\%             & 0.00\%             & 0.55\%          & 0.00      & 0.35                \\ 
                                  & LFW                      & 100.0\%              & 92.67\%             & 0.00\%             & 64.07\%           & 0.00     & 0.69                \\ 
                                  & Selfie                   & 41.36\%              & 11.05\%             & 0.00\%             & 6.37\%           & 0.00     & 0.57                \\ 
                                  & SFHQ                     & 100.0\%              & 100.0\%             & 0.00\%             & 99.73\%           &  0.00   & \textbf{1.00}                \\ 
                                  & SDFace                   & 86.89\%              & 99.39\%             & 0.00\%             & 64.50\%           &  0.00   & 0.65                \\ \midrule
        \multirow{5}{*}{ViT}     & CelebA                   & 79.76\%              & 23.61\%             & 0.00\%             & 2.24\%            & 0.00   & 0.09               \\ 
                                  & LFW                      &99.89\%              & 98.34\%             & 0.00\%             & 90.37\%           &  0.00   & 0.92                \\ 
                                  & Selfie                   & 10.11\%              & 49.85\%             & 0.00\%             & 12.77\%           & 0.00    & 0.25                \\ 
                                  & SFHQ                     & 100.0\%              &100.0\%             & 0.00\%             & 99.78\%           &  0.00   & \textbf{1.00}                \\ 
                                  & SDFace                   & 97.61\%              & 53.83\%             & 0.00\%             & 9.33\%           &  0.00   & 0.17                \\ \midrule
        \multirow{5}{*}{DIRE}    & CelebA                   & 91.82\%              &72.05\%         & 0.83\%                 & \textbf{28.00\%}          &  0.01     &\textbf{ 0.39}            \\ 
                                  & LFW                      & 99.96\%              & 99.96\%             &65.51\%             & \textbf{99.96\%}           & 0.65    &\textbf{1.00}                \\ 
                                  & Selfie                   &26.11\%              & 25.60\%             & 7.28\%             & \textbf{40.47\%}           & 0.28    & \textbf{1.58}                \\ 
                                  & SFHQ                     & 100.0\%              &100.0\%             &\textbf{100.0\%}             & \textbf{100.0\%}           &  \textbf{1.00}   &\textbf{ 1.00}                \\ 
                                  & SDFace                   & 92.83\%              &81.28\%             & \textbf{78.72\%}             & 41.06\%            & \textbf{0.85}   & 0.50                \\ 
       
        \bottomrule
    \end{tabular}
\end{table*}

\section {Experiments}

\subsection{Datasets}
In this study, we employ two well-established classification models, ResNet50~\cite{resnet} and Vision Transformer (ViT)~\cite{vit}, to detect AI-generated images. Each model is independently utilized in our framework to support detection tasks in facial analysis.
For training and evaluation, we leverage multiple datasets that encompass diverse facial characteristics and scenarios. The real face datasets used include CelebA~\cite{celeba}, LFW~\cite{lfw}, and Selfie~\cite{selfie}. For AI-generated faces, we select SDFace~\cite{sdface} and SFHQ~\cite{sfhq}.
Please refer to \textit{Appendix} for details of these datasets.

For images in different datasets, the clarity may vary; therefore, the images were converted to JPEG format with a quality setting of 95, and during pre-processing for the model, they were consistently cropped to a size of 224x224 pixels.\looseness=-1

\subsection{Implementation Details}

We deployed our method on an NVIDIA A100 GPU. The ResNet50 and ViT-B/16 are pre-trained on ImageNet and were employed as our backbone detectors. For each setting, Adam optimizer is used with a learning rate of 5e-6 and $\beta_1 = 0.9$, $\beta_2 = 0.999$. During training, we conducted 10 epochs per process with a batch size of 128. The trade-off parameter $\lambda=1$ of Eqn.~\ref{equ:loss}.

The attack method used in this study is Projected Gradient Descent (PGD), applied under the $L_\infty$  norm constraint with a step size of 10.
Training with the ResNet50 backbone took approximately 3.3 hours without adversarial training and 15 hours with adversarial training, while a single evaluation required about 9.5 ms.
Training with DIRE required approximately 9 hours to extract DIRE images, followed by 3.5 hours to train the classifier. When applying adversarial training with DIRE, the total computational cost was approximately doubled.


\subsection{Detection Performance Metrics}

Since testing on a single dataset may only produce binary outcomes (true or false), Precision and Recall metrics may lack significance. Therefore, we do not report Precision and Recall.
Furthermore, to systematically evaluate the resistance of models, we propose a robustness score, as follows:
\begin{equation}
    \text{Robustness Score} = \frac{\text{Acc}_{\text{adv}}}{\text{Acc}_{\text{clean}}},
\end{equation}
where $\text{Acc}_{\text{adv}}$ and $\text{Acc}_{\text{clean}}$ denotes accuracy under adversarial (attacked) and clean (unattacked) contexts respectively.
We will report both Accuracy and Robustness Score in the following subsections.




We adopt two evaluation settings in our experiments: 

\begin{itemize}[leftmargin=*]
    \item \textbf{All-set}: In this setting, detectors are trained on the combined datasets and tested on their respective test sets. This approach evaluates the overall performance of the detectors across all datasets. The results are in Table \ref{tab:single_model};
    \item \textbf{Cross-domain}: In this setting, detectors are trained on a selected subset of datasets (one real and one fake dataset) and tested on the remaining test sets from the other datasets. This approach assesses the model's generalization ability to unseen data distributions. The results of training on LFW and SFHQ while testing on other datasets are shown in Table \ref{tab:cross-model_lfw_sfhq}.
\end{itemize}
In these tables, we report results for both clean images and images with adversarial perturbations. Additionally, we compare models with adversarial training (``w/ AT'') and without adversarial training (``w/o AT''). 
For the rest of the combinations, we will report in the \textit{Appendix}.

\subsection{All-Set Detection Performance}

\subsubsection{Impact of Adversarial Perturbations.}

As shown in Table~\ref{tab:single_model}, comparing  ``Clean Images / w/o AT'' and ``Adversarial Images / w/o AT'' highlights the impact of adversarial perturbations. 
Under the standard setup without adversarial perturbations or adversarial training, test accuracy for both ResNet and ViT architectures approaches 100\%, demonstrating near-perfect performance.
This holds true for both the all-set and cross-dataset settings, indicating minimal challenges under ideal conditions.
However, when subtle white-box adversarial perturbations are introduced, test accuracy and robustness scores drop sharply to near 0, rendering the detector ineffective. Even DIRE, a method explicitly designed for detecting fake images generated by diffusion models, experiences significant performance degradation across all subsets.
That said, DIRE shows a certain level of resilience on specific datasets, such as SDFace, Selfie and CelebA, where its performance against adversarial attacks is comparatively better. These results illustrate the high vulnerability of detection models to adversarial attack techniques while highlighting DIRE's potential to mitigate such vulnerabilities under certain conditions. 

\begin{table*}[t]
    \centering
    \caption{Performance obtained through training on CelebA and SDFace using different methods.}
    \label{tab:cross-model_celeba_sdface}
    \begin{tabular}{@{}ccccccccc@{}}
        \toprule
        \multirow{2}{*}{Method} & \multirow{2}{*}{Dataset} & \multicolumn{2}{c}{Clean Images} & \multicolumn{2}{c}{Adversarial Images} &  \multicolumn{2}{c}{Robustness Score} \\ \cmidrule(lr){3-4} \cmidrule(lr){5-6} \cmidrule(lr){7-8}
                                  &                           & w/o AT     & w/ AT             & w/o AT        & w/ AT   & w/o AT & w/ AT  \\ \midrule
        \multirow{5}{*}{ResNet}  & CelebA                   & 100.0\%              & 99.98\%             & 0.00\%             & 99.56\%          & 0.00      & 0.99                \\ 
                                  & LFW                      & 100.0\%              & 95.96\%             & 0.00\%             & 21.23\%           & 0.00     & 0.22                \\ 
                                  & Selfie                   & 28.77\%              & 92.03\%             & 0.00\%             & \textbf{48.13\%}           & 0.00     & 0.52                \\ 
                                  & SFHQ                     & 100.0\%              & 1.50\%             & 0.00\%             & 0.00\%           &  0.00   & 0.00                \\ 
                                  & SDFace                   & 100.0\%              & 89.17\%             & 0.00\%             & 47.72\%           &  0.00   & 0.53                \\ \midrule
        \multirow{5}{*}{ViT}     & CelebA                   & 100.0\%              & 100.0\%             & 0.00\%             & 99.88\%            & 0.00   & \textbf{ 1.00}               \\ 
                                  & LFW                      &100.0\%              & 91.80\%             & 0.00\%             & 34.95\%           &  0.00   & 0.38                \\ 
                                  & Selfie                   & 23.84\%              & 67.74\%             & 0.00\%             & 26.65\%           & 0.00    & 0.39                \\ 
                                  & SFHQ                     & 98.44\%              &28.35\%             & 0.00\%             & 2.13\%           &  0.00   & 0.07                \\ 
                                  & SDFace                   & 100.0\%              & 97.33\%             & 0.00\%             & 85.44\%           &  0.00   & 0.87                \\ \midrule
        \multirow{5}{*}{DIRE}    & CelebA                   & 100.0\%              & 100.0\%         & 58.93\%                 & \textbf{100.0\%}          &  0.59     & \textbf{1.00}            \\ 
                                  & LFW                      & 98.07\%              & 97.81\%             & 64.11\%             & \textbf{90.03\%}           & 0.65    &\textbf{0.92}                \\ 
                                  & Selfie                   & 19.27\%              & 23.85\%             & 8.76\%             & 27.58\%           & 0.45    & \textbf{1.15}                \\ 
                                  & SFHQ                     & 97.52\%              &85.20\%             &\textbf{95.45\% }            & 66.42\%           &  \textbf{0.98}   & 0.78                \\ 
                                  & SDFace                   & 100.0\%              &99.80\%             & 99.56\%             & \textbf{99.78\%}            & 0.99   & \textbf{1.00}               \\ 
       
        \bottomrule
    \end{tabular}
\end{table*}

\subsubsection{Effectiveness of Adversarial Training.}

As shown in Tables~\ref{tab:single_model}, comparing ``Adversarial Images / w/o AT'' and ``Adversarial Images / w/ AT'' highlights the significant improvement in accuracy achieved through adversarial training. This improvement is attributed to the inclusion of hard samples during training, which helps the detector further reduce the error space beyond the standard setup, particularly under adversarial attack conditions.

When comparing ``Clean Images / w/ AT'' and ``Adversarial Images / w/ AT'', the performance remains comparable, demonstrating that adversarial training effectively preserves accuracy even under attack conditions. Experimental results highlight the importance of tailored adversarial training in enhancing the robustness of the detector, as reflected in the improved robustness scores.


\subsubsection{Effectiveness of DIRE}

We evaluate the performance of DIRE under both standard and perturbed conditions.
A comparison between DIRE, ResNet and ViT in the ``Adversarial Images / w/o AT'' columns highlights a significant improvement in accuracy achieved through DIRE. This enhancement can be attributed to DIRE’s ability to mitigate adversarial noise using its diffusion inversion and reconstruction process.
As shown in the ``Robustness Score'' column of Table~\ref{tab:single_model}, compared to ResNet and ViT, which directly operate in pixel space, DIRE exhibits a certain level of robustness against white-box attacks. This robustness stems from the inherent denoising effect of DIRE during image reconstruction using DDIM, which helps suppress adversarial noise and prevents a drastic decline in accuracy. However, this denoising effect is not absolute, as subtle perturbations can still degrade the performance of the detector.

\subsubsection{Visualization}
To analyze the anti-attack properties of DIRE, we visualized the DIRE maps of real and synthetic images before and after adversarial attacks, as shown in Fig. \ref{fig:cp} and Fig. \ref{fig:cp1}. For clarity, we computed the absolute point-wise difference between the DIRE maps before and after the attacks, amplified it by a factor of 10, and presented it in the last column.

DIRE maps of synthetic images appear lighter than those of real images, which aligns with expectations due to the DDIM inversion/reconstruction process. 
From the figures, we observe that the adversarial noise in synthetic images is smoother than in real images, but this effect is imperceptible in the original images.
Notably, compared to synthetic images, the difference maps of real images exhibit sharper, white-highlighted regions, primarily concentrated at the edges. 
This observation indicates that the discrepancy between the DIRE maps of real images after an attack and their original DIRE maps is more provoked than that of synthetic images. 
In other words, DIRE amplifies the differences between real and synthetic images under adversarial noise, revealing patterns that remain indistinguishable in the original image space.

With this observation, along with the results presented in the ``DIRE'' rows under the ``w/ AT'' column of Table~\ref{tab:single_model}, we confirm that this configuration achieves optimal performance under white-box attack conditions.
The results highlight two key points: first, training detection models in the DIRE space enhances the adversarial robustness; second, training based on DIRE, when combined with adversarial training, enables more precise discrimination by leveraging the amplified differences between real and synthetic images.\looseness=-1

\subsection{Cross-Domain Generalization Performance}

Rather than restricting training and testing to the same distribution, this section evaluates the generalization capability of our method across different domains. Specifically, we selected training sets from one real and one synthetic dataset and then tested on all five datasets. The evaluation results for training on LFW and SFHQ are presented in Table~\ref{tab:cross-model_lfw_sfhq}, training on CelebA and SDFace in Table~\ref{tab:cross-model_celeba_sdface}.
The results for other randomly selected training sets are provided in the \textit{Appendix}.
The performance of nearly all configurations declines significantly under cross-dataset testing scenarios, and adversarial attacks further exacerbate this issue, reducing the detector’s accuracy to 0\%. \looseness=-1

Incorporating adversarial training enhances cross-domain robustness. For instance, as shown in Table~\ref{tab:cross-model_lfw_sfhq}, even when SDFace is not included in the training set, the post-attack accuracy improves from 0\% to 64.5\%. 
This improvement can be attributed to the detector learning not only the inherent characteristics of real and synthetic images but also generalized representations of adversarial noise to new datasets.
Furthermore, leveraging DIRE further stabilizes the detector. A comparison between DIRE, ResNet and ViT in columns ``Adversarial Images / w/o AT'' columns highlights a significant improvement in accuracy achieved through DIRE, regardless of whether the dataset is trained or not. DIRE, with pre-trained diffusion models, maintains robustness under such conditions, allowing our method to achieve the highest resistance across most test sets.

A similar trend is observed in Table~\ref{tab:cross-model_celeba_sdface}.
Undoubtedly, the attack noise triggers the classifier's complete failure across all datasets for ResNet and ViT. 
However, incorporating adversarial samples into training significantly improves performance in both seen and unseen domains, although it does not fully restore accuracy to pre-attack levels.
Further integrating the diffusion reconstruction error further improves the accuracy, recovering performance to near pre-attack levels.
It is noteworthy that the accuracy of Selfie remains relatively low in both the standard setup (28.77\%) and the final setup with adversarial training and reconstruction error (27.58\%).
This may be due to the heavy stylization and post-processing applied to Selfie images, which likely introduced misclassifications.

Across all adversarial image results in Table~\ref{tab:cross-model_lfw_sfhq} and Table~\ref{tab:cross-model_celeba_sdface}, we observe that DIRE improves model performance in the ``w/o AT'' setting, demonstrating its ability to mitigate some adversarial perturbations. Additionally, models in the ``w/ AT'' setting always outperform those in the ``w/o AT'' setting, further validating the effectiveness of adversarial training in improving robustness.

\begin{figure}[t]
    \centering
    \includegraphics[width=1\linewidth]{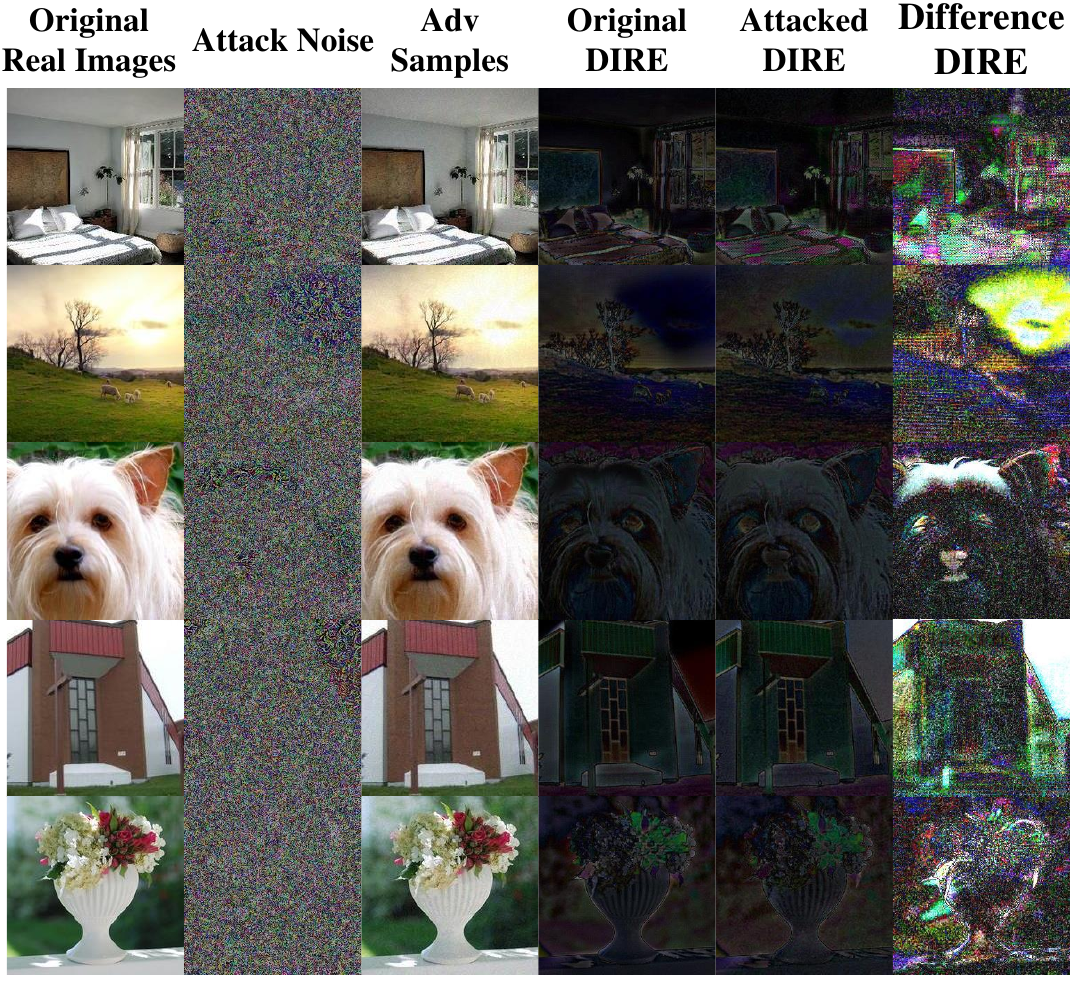} 
    \caption{Visualization of attack noise, adversarial samples, and DIRE maps of \textbf{non-face real images}.}
    \label{fig:normal_real}
\end{figure}

\vspace{-5pt}
\begin{table}[t]
    \centering
    \caption{Performance on non-face datasets.}
    \label{tab:limitation}
    \resizebox{1.0\linewidth}{!}{ 
    \begin{tabular}{@{}ccccccc@{}}
        \toprule
        \multirow{2}{*}{Method} & \multirow{2}{*}{Dataset} & \multicolumn{2}{c}{Clean Images} & \multicolumn{2}{c}{Adversarial Images}\\ \cmidrule(lr){3-4} \cmidrule(lr){5-6} 
                                  &                           & w/o AT     & w/ AT             & w/o AT        & w/ AT  \\ \midrule
        \multirow{2}{*}{ResNet}  & DF-imagenet                   & 99.28\%              & 67.62\%             & 0.00\%             & 67.62\%                    \\ 
                                  & Artifact                      & 100.0\%              & 95.96\%             & 0.00\%             & 21.23\%                       \\  \midrule
        
        \multirow{2}{*}{DIRE}    & DF-imagenet                   & 99.30\%              & 98.65\%         & 74.86\%                 &98.05\%              \\ 
                                  & Artifact                     & 56.61\%              & 57.04\%             & 54.23\%             & 99.40\%                    \\

        \bottomrule
    \end{tabular}
    }
    \label{tab:accuracy}

\end{table}
\vspace{-5pt}

\begin{figure}[t]
    \centering
    \includegraphics[width=1\linewidth]{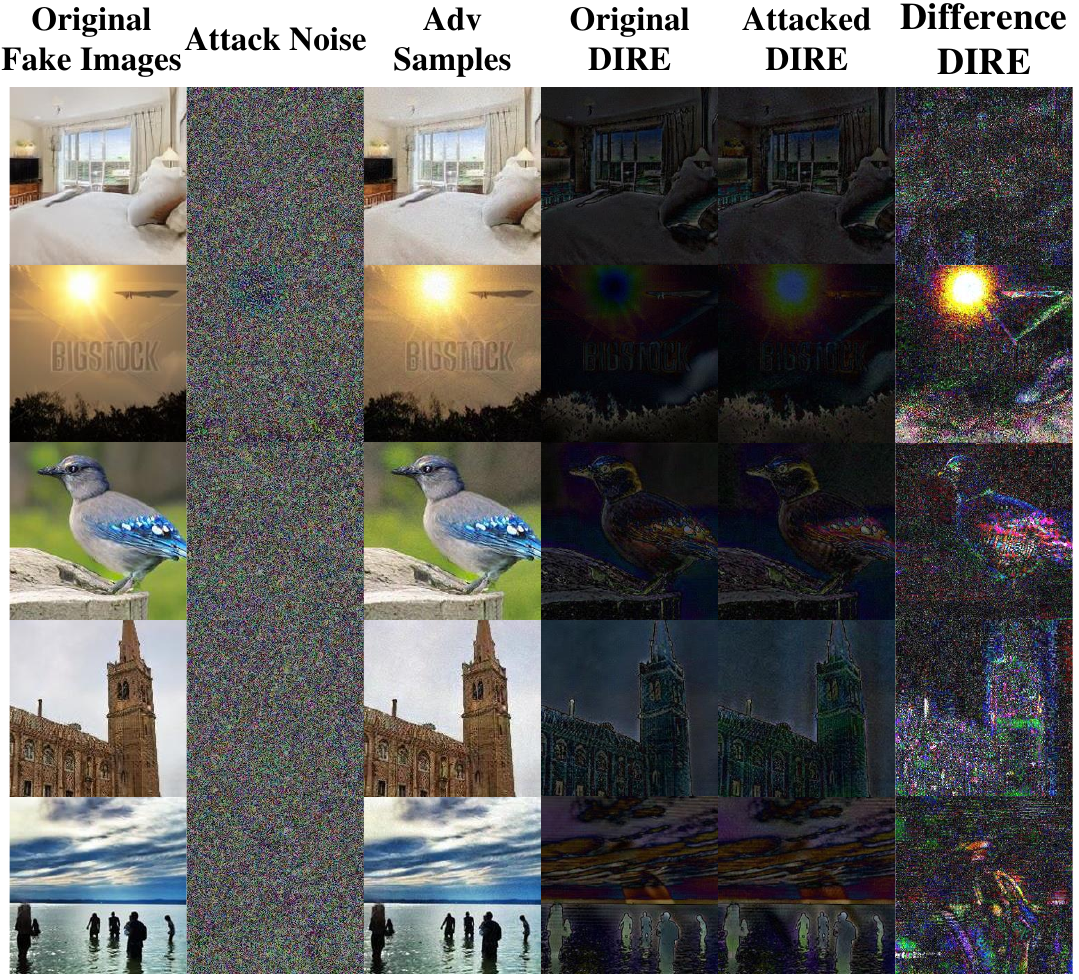} 
    \caption{Visualization of attack noise, adversarial samples, and DIRE maps of \textbf{non-face fake images}.}
    \label{fig:normal_fake}
\end{figure}

\section{Discussions}

In this paper, we primarily focus on face images. However, we also conduct experiments on natural images by employing Diffusion Forensics~\cite{dire} (DF-Imagenet) and the Artifact~\cite{artifact}. We observed that adversarial training caused the detection model to fail when tested on fake images. This failure manifested as fixed output predictions (either always true or always false), as shown in Table~\ref{tab:limitation}.
To further analyze this issue, we visualize adversarial noise, DIRE, and their differences in Figs.~\ref{fig:normal_real} and \ref{fig:normal_fake}. Unlike facial images, adversarial noise in non-face images is less smooth. Additionally, the DIRE differences for both real and fake samples are sharp, diminishing the amplification effect of DIRE. 
Our analysis suggests that this issue arises due to fundamental differences in image patterns and complexity. In face datasets, the relatively fixed patterns make it easier to identify inconsistencies in AI-generated images. However, ImageNet encompasses a diverse range of images with complex backgrounds, varied object types, lighting conditions, and pose variations, making detection models more susceptible to these factors.
When adversarial samples are introduced during training on such diverse datasets, the model struggles to learn sufficient universal features, leading to poor generalization and degraded performance.

\section{Conclusions}
Our research demonstrates that current generative face image detection models are highly susceptible to malicious perturbations that are imperceptible to humans yet significantly degrade model accuracy. By incorporating adversarial training, we can substantially enhance the model's resilience to adversarial challenges. Diffusion reconstruction further provides a promising approach to improving robustness.
We evaluate our method in both in-domain and cross-domain settings to comprehensively assess its effectiveness in enhancing detection robustness.

\bibliographystyle{ACM-Reference-Format}
\bibliography{main}


\begin{thebibliography}{44}


\ifx \showCODEN    \undefined \def \showCODEN     #1{\unskip}     \fi
\ifx \showISBNx    \undefined \def \showISBNx     #1{\unskip}     \fi
\ifx \showISBNxiii \undefined \def \showISBNxiii  #1{\unskip}     \fi
\ifx \showISSN     \undefined \def \showISSN      #1{\unskip}     \fi
\ifx \showLCCN     \undefined \def \showLCCN      #1{\unskip}     \fi
\ifx \shownote     \undefined \def \shownote      #1{#1}          \fi
\ifx \showarticletitle \undefined \def \showarticletitle #1{#1}   \fi
\ifx \showURL      \undefined \def \showURL       {\relax}        \fi
\providecommand\bibfield[2]{#2}
\providecommand\bibinfo[2]{#2}
\providecommand\natexlab[1]{#1}
\providecommand\showeprint[2][]{arXiv:#2}

\bibitem[Baraldi et~al\mbox{.}(2025)]%
        {code}
\bibfield{author}{\bibinfo{person}{Lorenzo Baraldi}, \bibinfo{person}{Federico Cocchi}, \bibinfo{person}{Marcella Cornia}, \bibinfo{person}{Alessandro Nicolosi}, {and} \bibinfo{person}{Rita Cucchiara}.} \bibinfo{year}{2025}\natexlab{}.
\newblock \showarticletitle{Contrasting deepfakes diffusion via contrastive learning and global-local similarities}. In \bibinfo{booktitle}{\emph{European Conference on Computer Vision}}. Springer, \bibinfo{pages}{199--216}.
\newblock


\bibitem[Beniaguev(2022)]%
        {sfhq}
\bibfield{author}{\bibinfo{person}{David Beniaguev}.} \bibinfo{year}{2022}\natexlab{}.
\newblock \bibinfo{title}{Synthetic Faces High Quality (SFHQ) dataset}.
\newblock
\href{https://doi.org/10.34740/kaggle/dsv/4737549}{doi:\nolinkurl{10.34740/kaggle/dsv/4737549}}


\bibitem[Borji(2023)]%
        {185}
\bibfield{author}{\bibinfo{person}{Ali Borji}.} \bibinfo{year}{2023}\natexlab{}.
\newblock \showarticletitle{Qualitative failures of image generation models and their application in detecting deepfakes}.
\newblock \bibinfo{journal}{\emph{Image and Vision Computing}}  \bibinfo{volume}{137} (\bibinfo{year}{2023}), \bibinfo{pages}{104771}.
\newblock


\bibitem[Chen et~al\mbox{.}(2024)]%
        {drct}
\bibfield{author}{\bibinfo{person}{Baoying Chen}, \bibinfo{person}{Jishen Zeng}, \bibinfo{person}{Jianquan Yang}, {and} \bibinfo{person}{Rui Yang}.} \bibinfo{year}{2024}\natexlab{}.
\newblock \showarticletitle{DRCT: Diffusion Reconstruction Contrastive Training towards Universal Detection of Diffusion Generated Images}. In \bibinfo{booktitle}{\emph{Forty-first International Conference on Machine Learning}}.
\newblock


\bibitem[Choi et~al\mbox{.}(2024)]%
        {v1}
\bibfield{author}{\bibinfo{person}{Jongwook Choi}, \bibinfo{person}{Taehoon Kim}, \bibinfo{person}{Yonghyun Jeong}, \bibinfo{person}{Seungryul Baek}, {and} \bibinfo{person}{Jongwon Choi}.} \bibinfo{year}{2024}\natexlab{}.
\newblock \showarticletitle{Exploiting Style Latent Flows for Generalizing Deepfake Video Detection}. In \bibinfo{booktitle}{\emph{Proceedings of the IEEE/CVF Conference on Computer Vision and Pattern Recognition}}. \bibinfo{pages}{1133--1143}.
\newblock


\bibitem[Croce and Hein(2020)]%
        {autoattack}
\bibfield{author}{\bibinfo{person}{Francesco Croce} {and} \bibinfo{person}{Matthias Hein}.} \bibinfo{year}{2020}\natexlab{}.
\newblock \showarticletitle{Reliable evaluation of adversarial robustness with an ensemble of diverse parameter-free attacks}. In \bibinfo{booktitle}{\emph{International conference on machine learning}}. PMLR, \bibinfo{pages}{2206--2216}.
\newblock


\bibitem[Dhariwal and Nichol(2021)]%
        {dmbgi}
\bibfield{author}{\bibinfo{person}{Prafulla Dhariwal} {and} \bibinfo{person}{Alexander Nichol}.} \bibinfo{year}{2021}\natexlab{}.
\newblock \showarticletitle{Diffusion models beat gans on image synthesis}.
\newblock \bibinfo{journal}{\emph{Advances in neural information processing systems}}  \bibinfo{volume}{34} (\bibinfo{year}{2021}), \bibinfo{pages}{8780--8794}.
\newblock


\bibitem[Dosovitskiy(2020)]%
        {vit}
\bibfield{author}{\bibinfo{person}{Alexey Dosovitskiy}.} \bibinfo{year}{2020}\natexlab{}.
\newblock \showarticletitle{An image is worth 16x16 words: Transformers for image recognition at scale}.
\newblock \bibinfo{journal}{\emph{arXiv preprint arXiv:2010.11929}} (\bibinfo{year}{2020}).
\newblock


\bibitem[Farid(2022a)]%
        {187}
\bibfield{author}{\bibinfo{person}{Hany Farid}.} \bibinfo{year}{2022}\natexlab{a}.
\newblock \showarticletitle{Lighting (in) consistency of paint by text}.
\newblock \bibinfo{journal}{\emph{arXiv preprint arXiv:2207.13744}} (\bibinfo{year}{2022}).
\newblock


\bibitem[Farid(2022b)]%
        {186}
\bibfield{author}{\bibinfo{person}{Hany Farid}.} \bibinfo{year}{2022}\natexlab{b}.
\newblock \showarticletitle{Perspective (in) consistency of paint by text}.
\newblock \bibinfo{journal}{\emph{arXiv preprint arXiv:2206.14617}} (\bibinfo{year}{2022}).
\newblock


\bibitem[Goodfellow et~al\mbox{.}(2014a)]%
        {gan}
\bibfield{author}{\bibinfo{person}{Ian Goodfellow}, \bibinfo{person}{Jean Pouget-Abadie}, \bibinfo{person}{Mehdi Mirza}, \bibinfo{person}{Bing Xu}, \bibinfo{person}{David Warde-Farley}, \bibinfo{person}{Sherjil Ozair}, \bibinfo{person}{Aaron Courville}, {and} \bibinfo{person}{Yoshua Bengio}.} \bibinfo{year}{2014}\natexlab{a}.
\newblock \showarticletitle{Generative adversarial nets}.
\newblock \bibinfo{journal}{\emph{Advances in neural information processing systems}}  \bibinfo{volume}{27} (\bibinfo{year}{2014}).
\newblock


\bibitem[Goodfellow et~al\mbox{.}(2014b)]%
        {fgsm}
\bibfield{author}{\bibinfo{person}{Ian~J Goodfellow}, \bibinfo{person}{Jonathon Shlens}, {and} \bibinfo{person}{Christian Szegedy}.} \bibinfo{year}{2014}\natexlab{b}.
\newblock \showarticletitle{Explaining and harnessing adversarial examples}.
\newblock \bibinfo{journal}{\emph{arXiv preprint arXiv:1412.6572}} (\bibinfo{year}{2014}).
\newblock


\bibitem[Goswami et~al\mbox{.}(2019)]%
        {advdefence3}
\bibfield{author}{\bibinfo{person}{Gaurav Goswami}, \bibinfo{person}{Akshay Agarwal}, \bibinfo{person}{Nalini Ratha}, \bibinfo{person}{Richa Singh}, {and} \bibinfo{person}{Mayank Vatsa}.} \bibinfo{year}{2019}\natexlab{}.
\newblock \showarticletitle{Detecting and mitigating adversarial perturbations for robust face recognition}.
\newblock \bibinfo{journal}{\emph{International Journal of Computer Vision}}  \bibinfo{volume}{127} (\bibinfo{year}{2019}), \bibinfo{pages}{719--742}.
\newblock


\bibitem[Goswami et~al\mbox{.}(2018)]%
        {advdefence2}
\bibfield{author}{\bibinfo{person}{Gaurav Goswami}, \bibinfo{person}{Nalini Ratha}, \bibinfo{person}{Akshay Agarwal}, \bibinfo{person}{Richa Singh}, {and} \bibinfo{person}{Mayank Vatsa}.} \bibinfo{year}{2018}\natexlab{}.
\newblock \showarticletitle{Unravelling robustness of deep learning based face recognition against adversarial attacks}. In \bibinfo{booktitle}{\emph{Proceedings of the AAAI Conference on Artificial Intelligence}}, Vol.~\bibinfo{volume}{32}.
\newblock


\bibitem[He et~al\mbox{.}(2016)]%
        {resnet}
\bibfield{author}{\bibinfo{person}{Kaiming He}, \bibinfo{person}{Xiangyu Zhang}, \bibinfo{person}{Shaoqing Ren}, {and} \bibinfo{person}{Jian Sun}.} \bibinfo{year}{2016}\natexlab{}.
\newblock \showarticletitle{Deep residual learning for image recognition}. In \bibinfo{booktitle}{\emph{Proceedings of the IEEE conference on computer vision and pattern recognition}}. \bibinfo{pages}{770--778}.
\newblock


\bibitem[Ho et~al\mbox{.}(2020)]%
        {ddpm}
\bibfield{author}{\bibinfo{person}{Jonathan Ho}, \bibinfo{person}{Ajay Jain}, {and} \bibinfo{person}{Pieter Abbeel}.} \bibinfo{year}{2020}\natexlab{}.
\newblock \showarticletitle{Denoising diffusion probabilistic models}.
\newblock \bibinfo{journal}{\emph{Advances in neural information processing systems}}  \bibinfo{volume}{33} (\bibinfo{year}{2020}), \bibinfo{pages}{6840--6851}.
\newblock


\bibitem[Huang et~al\mbox{.}(2008)]%
        {lfw}
\bibfield{author}{\bibinfo{person}{Gary~B Huang}, \bibinfo{person}{Marwan Mattar}, \bibinfo{person}{Tamara Berg}, {and} \bibinfo{person}{Eric Learned-Miller}.} \bibinfo{year}{2008}\natexlab{}.
\newblock \showarticletitle{Labeled faces in the wild: A database forstudying face recognition in unconstrained environments}. In \bibinfo{booktitle}{\emph{Workshop on faces in'Real-Life'Images: detection, alignment, and recognition}}.
\newblock


\bibitem[Kalayeh et~al\mbox{.}(2015)]%
        {selfie}
\bibfield{author}{\bibinfo{person}{Mahdi~M Kalayeh}, \bibinfo{person}{Misrak Seifu}, \bibinfo{person}{Wesna LaLanne}, {and} \bibinfo{person}{Mubarak Shah}.} \bibinfo{year}{2015}\natexlab{}.
\newblock \showarticletitle{How to take a good selfie?}. In \bibinfo{booktitle}{\emph{Proceedings of the 23rd ACM international conference on Multimedia}}. \bibinfo{pages}{923--926}.
\newblock


\bibitem[Karras(2019)]%
        {karras2019style}
\bibfield{author}{\bibinfo{person}{Tero Karras}.} \bibinfo{year}{2019}\natexlab{}.
\newblock \showarticletitle{A Style-Based Generator Architecture for Generative Adversarial Networks}.
\newblock \bibinfo{journal}{\emph{arXiv preprint arXiv:1812.04948}} (\bibinfo{year}{2019}).
\newblock


\bibitem[Karras et~al\mbox{.}(2019)]%
        {stylegan}
\bibfield{author}{\bibinfo{person}{Tero Karras}, \bibinfo{person}{Samuli Laine}, {and} \bibinfo{person}{Timo Aila}.} \bibinfo{year}{2019}\natexlab{}.
\newblock \showarticletitle{A style-based generator architecture for generative adversarial networks}. In \bibinfo{booktitle}{\emph{Proceedings of the IEEE/CVF conference on computer vision and pattern recognition}}. \bibinfo{pages}{4401--4410}.
\newblock


\bibitem[Liu et~al\mbox{.}(2015)]%
        {celeba}
\bibfield{author}{\bibinfo{person}{Ziwei Liu}, \bibinfo{person}{Ping Luo}, \bibinfo{person}{Xiaogang Wang}, {and} \bibinfo{person}{Xiaoou Tang}.} \bibinfo{year}{2015}\natexlab{}.
\newblock \showarticletitle{Deep Learning Face Attributes in the Wild}. In \bibinfo{booktitle}{\emph{Proceedings of International Conference on Computer Vision (ICCV)}}.
\newblock


\bibitem[Ma et~al\mbox{.}(2023)]%
        {184}
\bibfield{author}{\bibinfo{person}{Ruipeng Ma}, \bibinfo{person}{Jinhao Duan}, \bibinfo{person}{Fei Kong}, \bibinfo{person}{Xiaoshuang Shi}, {and} \bibinfo{person}{Kaidi Xu}.} \bibinfo{year}{2023}\natexlab{}.
\newblock \showarticletitle{Exposing the fake: Effective diffusion-generated images detection}.
\newblock \bibinfo{journal}{\emph{arXiv preprint arXiv:2307.06272}} (\bibinfo{year}{2023}).
\newblock


\bibitem[Madry(2017)]%
        {pgd}
\bibfield{author}{\bibinfo{person}{Aleksander Madry}.} \bibinfo{year}{2017}\natexlab{}.
\newblock \showarticletitle{Towards deep learning models resistant to adversarial attacks}.
\newblock \bibinfo{journal}{\emph{arXiv preprint arXiv:1706.06083}} (\bibinfo{year}{2017}).
\newblock


\bibitem[Miyato et~al\mbox{.}(2016)]%
        {fgm}
\bibfield{author}{\bibinfo{person}{Takeru Miyato}, \bibinfo{person}{Andrew~M Dai}, {and} \bibinfo{person}{Ian Goodfellow}.} \bibinfo{year}{2016}\natexlab{}.
\newblock \showarticletitle{Adversarial training methods for semi-supervised text classification}.
\newblock \bibinfo{journal}{\emph{arXiv preprint arXiv:1605.07725}} (\bibinfo{year}{2016}).
\newblock


\bibitem[Nichol and Dhariwal(2021)]%
        {iddpm}
\bibfield{author}{\bibinfo{person}{Alexander~Quinn Nichol} {and} \bibinfo{person}{Prafulla Dhariwal}.} \bibinfo{year}{2021}\natexlab{}.
\newblock \showarticletitle{Improved denoising diffusion probabilistic models}. In \bibinfo{booktitle}{\emph{International conference on machine learning}}. PMLR, \bibinfo{pages}{8162--8171}.
\newblock


\bibitem[Nie et~al\mbox{.}(2022)]%
        {review1_4}
\bibfield{author}{\bibinfo{person}{Weili Nie}, \bibinfo{person}{Brandon Guo}, \bibinfo{person}{Yujia Huang}, \bibinfo{person}{Chaowei Xiao}, \bibinfo{person}{Arash Vahdat}, {and} \bibinfo{person}{Anima Anandkumar}.} \bibinfo{year}{2022}\natexlab{}.
\newblock \showarticletitle{Diffusion models for adversarial purification}.
\newblock \bibinfo{journal}{\emph{arXiv preprint arXiv:2205.07460}} (\bibinfo{year}{2022}).
\newblock


\bibitem[Oorloff et~al\mbox{.}(2024)]%
        {v2}
\bibfield{author}{\bibinfo{person}{Trevine Oorloff}, \bibinfo{person}{Surya Koppisetti}, \bibinfo{person}{Nicol{\`o} Bonettini}, \bibinfo{person}{Divyaraj Solanki}, \bibinfo{person}{Ben Colman}, \bibinfo{person}{Yaser Yacoob}, \bibinfo{person}{Ali Shahriyari}, {and} \bibinfo{person}{Gaurav Bharaj}.} \bibinfo{year}{2024}\natexlab{}.
\newblock \showarticletitle{AVFF: Audio-Visual Feature Fusion for Video Deepfake Detection}. In \bibinfo{booktitle}{\emph{Proceedings of the IEEE/CVF Conference on Computer Vision and Pattern Recognition}}. \bibinfo{pages}{27102--27112}.
\newblock


\bibitem[Peebles and Xie(2023)]%
        {dit}
\bibfield{author}{\bibinfo{person}{William Peebles} {and} \bibinfo{person}{Saining Xie}.} \bibinfo{year}{2023}\natexlab{}.
\newblock \showarticletitle{Scalable diffusion models with transformers}. In \bibinfo{booktitle}{\emph{Proceedings of the IEEE/CVF International Conference on Computer Vision}}. \bibinfo{pages}{4195--4205}.
\newblock


\bibitem[Rahman et~al\mbox{.}(2023)]%
        {artifact}
\bibfield{author}{\bibinfo{person}{Md~Awsafur Rahman}, \bibinfo{person}{Bishmoy Paul}, \bibinfo{person}{Najibul~Haque Sarker}, \bibinfo{person}{Zaber Ibn~Abdul Hakim}, {and} \bibinfo{person}{Shaikh~Anowarul Fattah}.} \bibinfo{year}{2023}\natexlab{}.
\newblock \showarticletitle{Artifact: A large-scale dataset with artificial and factual images for generalizable and robust synthetic image detection}. In \bibinfo{booktitle}{\emph{2023 IEEE International Conference on Image Processing (ICIP)}}. IEEE, \bibinfo{pages}{2200--2204}.
\newblock


\bibitem[Ramesh et~al\mbox{.}(2022)]%
        {htigcl}
\bibfield{author}{\bibinfo{person}{Aditya Ramesh}, \bibinfo{person}{Prafulla Dhariwal}, \bibinfo{person}{Alex Nichol}, \bibinfo{person}{Casey Chu}, {and} \bibinfo{person}{Mark Chen}.} \bibinfo{year}{2022}\natexlab{}.
\newblock \showarticletitle{Hierarchical text-conditional image generation with clip latents}.
\newblock \bibinfo{journal}{\emph{arXiv preprint arXiv:2204.06125}} \bibinfo{volume}{1}, \bibinfo{number}{2} (\bibinfo{year}{2022}), \bibinfo{pages}{3}.
\newblock


\bibitem[Ren et~al\mbox{.}(2022)]%
        {advdefence1}
\bibfield{author}{\bibinfo{person}{Min Ren}, \bibinfo{person}{Yuhao Zhu}, \bibinfo{person}{Yunlong Wang}, {and} \bibinfo{person}{Zhenan Sun}.} \bibinfo{year}{2022}\natexlab{}.
\newblock \showarticletitle{Perturbation inactivation based adversarial defense for face recognition}.
\newblock \bibinfo{journal}{\emph{IEEE Transactions on Information Forensics and Security}}  \bibinfo{volume}{17} (\bibinfo{year}{2022}), \bibinfo{pages}{2947--2962}.
\newblock


\bibitem[Rombach et~al\mbox{.}(2022)]%
        {ldm}
\bibfield{author}{\bibinfo{person}{Robin Rombach}, \bibinfo{person}{Andreas Blattmann}, \bibinfo{person}{Dominik Lorenz}, \bibinfo{person}{Patrick Esser}, {and} \bibinfo{person}{Bj{\"o}rn Ommer}.} \bibinfo{year}{2022}\natexlab{}.
\newblock \showarticletitle{High-resolution image synthesis with latent diffusion models}. In \bibinfo{booktitle}{\emph{Proceedings of the IEEE/CVF conference on computer vision and pattern recognition}}. \bibinfo{pages}{10684--10695}.
\newblock


\bibitem[Sinitsa and Fried(2024)]%
        {102}
\bibfield{author}{\bibinfo{person}{Sergey Sinitsa} {and} \bibinfo{person}{Ohad Fried}.} \bibinfo{year}{2024}\natexlab{}.
\newblock \bibinfo{title}{Deep Image Fingerprint: Towards Low Budget Synthetic Image Detection and Model Lineage Analysis}.
\newblock
\showeprint[arxiv]{2303.10762}~[cs.CV]
\urldef\tempurl%
\url{https://arxiv.org/abs/2303.10762}
\showURL{%
\tempurl}


\bibitem[Song et~al\mbox{.}(2020)]%
        {ddim}
\bibfield{author}{\bibinfo{person}{Jiaming Song}, \bibinfo{person}{Chenlin Meng}, {and} \bibinfo{person}{Stefano Ermon}.} \bibinfo{year}{2020}\natexlab{}.
\newblock \showarticletitle{Denoising diffusion implicit models}.
\newblock \bibinfo{journal}{\emph{arXiv preprint arXiv:2010.02502}} (\bibinfo{year}{2020}).
\newblock


\bibitem[Tan et~al\mbox{.}(2024)]%
        {rethinking}
\bibfield{author}{\bibinfo{person}{Chuangchuang Tan}, \bibinfo{person}{Yao Zhao}, \bibinfo{person}{Shikui Wei}, \bibinfo{person}{Guanghua Gu}, \bibinfo{person}{Ping Liu}, {and} \bibinfo{person}{Yunchao Wei}.} \bibinfo{year}{2024}\natexlab{}.
\newblock \showarticletitle{Rethinking the up-sampling operations in cnn-based generative network for generalizable deepfake detection}. In \bibinfo{booktitle}{\emph{Proceedings of the IEEE/CVF Conference on Computer Vision and Pattern Recognition}}. \bibinfo{pages}{28130--28139}.
\newblock


\bibitem[Tian et~al\mbox{.}(2024)]%
        {var}
\bibfield{author}{\bibinfo{person}{Keyu Tian}, \bibinfo{person}{Yi Jiang}, \bibinfo{person}{Zehuan Yuan}, \bibinfo{person}{Bingyue Peng}, {and} \bibinfo{person}{Liwei Wang}.} \bibinfo{year}{2024}\natexlab{}.
\newblock \showarticletitle{Visual autoregressive modeling: Scalable image generation via next-scale prediction}.
\newblock \bibinfo{journal}{\emph{arXiv preprint arXiv:2404.02905}} (\bibinfo{year}{2024}).
\newblock


\bibitem[tobecwb(2023)]%
        {sdface}
\bibfield{author}{\bibinfo{person}{tobecwb}.} \bibinfo{year}{2023}\natexlab{}.
\newblock \bibinfo{title}{stable-diffusion-face-dataset}.
\newblock \bibinfo{howpublished}{\url{https://github.com/tobecwb/stable-diffusion-face-dataset}}.
\newblock


\bibitem[Wang et~al\mbox{.}(2023a)]%
        {dire}
\bibfield{author}{\bibinfo{person}{Zhendong Wang}, \bibinfo{person}{Jianmin Bao}, \bibinfo{person}{Wengang Zhou}, \bibinfo{person}{Weilun Wang}, \bibinfo{person}{Hezhen Hu}, \bibinfo{person}{Hong Chen}, {and} \bibinfo{person}{Houqiang Li}.} \bibinfo{year}{2023}\natexlab{a}.
\newblock \showarticletitle{Dire for diffusion-generated image detection}. In \bibinfo{booktitle}{\emph{Proceedings of the IEEE/CVF International Conference on Computer Vision}}. \bibinfo{pages}{22445--22455}.
\newblock


\bibitem[Wang et~al\mbox{.}(2023b)]%
        {review1_5}
\bibfield{author}{\bibinfo{person}{Zekai Wang}, \bibinfo{person}{Tianyu Pang}, \bibinfo{person}{Chao Du}, \bibinfo{person}{Min Lin}, \bibinfo{person}{Weiwei Liu}, {and} \bibinfo{person}{Shuicheng Yan}.} \bibinfo{year}{2023}\natexlab{b}.
\newblock \showarticletitle{Better diffusion models further improve adversarial training}. In \bibinfo{booktitle}{\emph{International conference on machine learning}}. PMLR, \bibinfo{pages}{36246--36263}.
\newblock


\bibitem[Wu et~al\mbox{.}(2023)]%
        {Lasted}
\bibfield{author}{\bibinfo{person}{Haiwei Wu}, \bibinfo{person}{Jiantao Zhou}, {and} \bibinfo{person}{Shile Zhang}.} \bibinfo{year}{2023}\natexlab{}.
\newblock \showarticletitle{Generalizable synthetic image detection via language-guided contrastive learning}.
\newblock \bibinfo{journal}{\emph{arXiv preprint arXiv:2305.13800}} (\bibinfo{year}{2023}).
\newblock


\bibitem[Xu et~al\mbox{.}(2024)]%
        {xu2024analyzing}
\bibfield{author}{\bibinfo{person}{Ying Xu}, \bibinfo{person}{Philipp Terh{\"o}st}, \bibinfo{person}{Marius Pedersen}, {and} \bibinfo{person}{Kiran Raja}.} \bibinfo{year}{2024}\natexlab{}.
\newblock \showarticletitle{Analyzing Fairness in Deepfake Detection With Massively Annotated Databases}.
\newblock \bibinfo{journal}{\emph{IEEE Transactions on Technology and Society}} (\bibinfo{year}{2024}).
\newblock


\bibitem[Yan et~al\mbox{.}(2024)]%
        {lsda}
\bibfield{author}{\bibinfo{person}{Zhiyuan Yan}, \bibinfo{person}{Yuhao Luo}, \bibinfo{person}{Siwei Lyu}, \bibinfo{person}{Qingshan Liu}, {and} \bibinfo{person}{Baoyuan Wu}.} \bibinfo{year}{2024}\natexlab{}.
\newblock \showarticletitle{Transcending forgery specificity with latent space augmentation for generalizable deepfake detection}. In \bibinfo{booktitle}{\emph{Proceedings of the IEEE/CVF Conference on Computer Vision and Pattern Recognition}}. \bibinfo{pages}{8984--8994}.
\newblock


\bibitem[Zhang et~al\mbox{.}(2025)]%
        {v3}
\bibfield{author}{\bibinfo{person}{Daichi Zhang}, \bibinfo{person}{Zihao Xiao}, \bibinfo{person}{Shikun Li}, \bibinfo{person}{Fanzhao Lin}, \bibinfo{person}{Jianmin Li}, {and} \bibinfo{person}{Shiming Ge}.} \bibinfo{year}{2025}\natexlab{}.
\newblock \showarticletitle{Learning natural consistency representation for face forgery video detection}. In \bibinfo{booktitle}{\emph{European Conference on Computer Vision}}. Springer, \bibinfo{pages}{407--424}.
\newblock


\bibitem[Zhu et~al\mbox{.}(2017)]%
        {cyclegan}
\bibfield{author}{\bibinfo{person}{Jun-Yan Zhu}, \bibinfo{person}{Taesung Park}, \bibinfo{person}{Phillip Isola}, {and} \bibinfo{person}{Alexei~A Efros}.} \bibinfo{year}{2017}\natexlab{}.
\newblock \showarticletitle{Unpaired image-to-image translation using cycle-consistent adversarial networks}. In \bibinfo{booktitle}{\emph{Proceedings of the IEEE international conference on computer vision}}. \bibinfo{pages}{2223--2232}.
\newblock


\end{thebibliography}

\end{document}